\documentclass{article}

\usepackage[preprint,nonatbib]{neurips_2024}

\usepackage[utf8]{inputenc} %
\usepackage[T1]{fontenc}    %
\usepackage{hyperref}       %
\usepackage{url}            %
\usepackage{booktabs}       %
\usepackage{amsfonts}       %
\usepackage{nicefrac}       %
\usepackage{microtype}      %
\usepackage{xcolor}         %

\usepackage[utf8]{inputenc} %
\usepackage[T1]{fontenc}    %
\usepackage{hyperref}       %
\usepackage{url}            %
\usepackage{booktabs}       %
\usepackage{amsfonts}       %
\usepackage{nicefrac}       %
\usepackage{microtype}      %
\usepackage{xcolor}         %

\usepackage{graphicx}
\usepackage{amsmath}
\usepackage{amssymb}
\usepackage{booktabs}
\usepackage{pifont}%
\usepackage{arydshln}
\usepackage{dblfloatfix}

\usepackage{colortbl} 
\usepackage{multirow}

\usepackage{caption}
\usepackage{wrapfig}

\usepackage{tikz}
\usetikzlibrary{bayesnet}
\usetikzlibrary{arrows}
\usepackage{float}

\usepackage[accsupp]{axessibility}
\usepackage{hyperref}       %

\usepackage[capitalize]{cleveref}
\crefname{section}{Sec.}{Secs.}
\Crefname{section}{Section}{Sections}
\Crefname{table}{Table}{Tables}
\crefname{table}{Tab.}{Tabs.}

\title{ED-SAM: An Efficient Diffusion Sampling Approach to Domain Generalization in Vision-Language Foundation Models}

\author{%
Thanh-Dat Truong$^{1}$, 
Xin Li$^{2}$, 
Bhiksha Raj$^{3,4}$, 
Jackson Cothren$^5$, 
Khoa Luu$^{1}$\\
$^{1}$CVIU Lab, University of Arkansas, USA \quad
$^{2}$University at Albany, Albany NY, USA  \\
$^{3}$Carnegie Mellon University, USA  \quad
$^{4}$Mohammed bin Zayed University of AI, UAE\\
$^{5}$Dep. of Geosciences, University of Arkansas, USA \\
\tt\small \{tt032,  jcothre, khoaluu\}@uark.edu \tt\small bhiksha@cs.cmu.edu, xli48@albany.edu
\vspace{-6mm}
}

\begin{document}

\maketitle

\begin{abstract}

The Vision-Language Foundation Model has recently shown outstanding performance in various perception learning tasks. The outstanding performance of the vision-language model mainly relies on large-scale pre-training datasets and different data augmentation techniques. However, the domain generalization problem of the vision-language foundation model needs to be addressed. This problem has limited the generalizability of the vision-language foundation model to unknown data distributions. In this paper, we introduce a new simple but efficient Diffusion Sampling approach to Domain Generalization (ED-SAM) to improve the generalizability of the vision-language foundation model. Our theoretical analysis in this work reveals the critical role and relation of the diffusion model to domain generalization in the vision-language foundation model. Then, based on the insightful analysis, we introduce a new simple yet effective Transport Transformation to diffusion sampling method. It can effectively generate adversarial samples to improve the generalizability of the foundation model against unknown data distributions. The experimental results on different scales of vision-language pre-training datasets, including CC3M, CC12M, and LAION400M, have consistently shown State-of-the-Art performance and scalability of the proposed ED-SAM approach compared to the other recent methods.
\end{abstract}

\section{Introduction}

\begin{wrapfigure}[17]{r}{0.5\textwidth}
    \centering
    \vspace{-8mm}
    \includegraphics[width=0.5\textwidth]{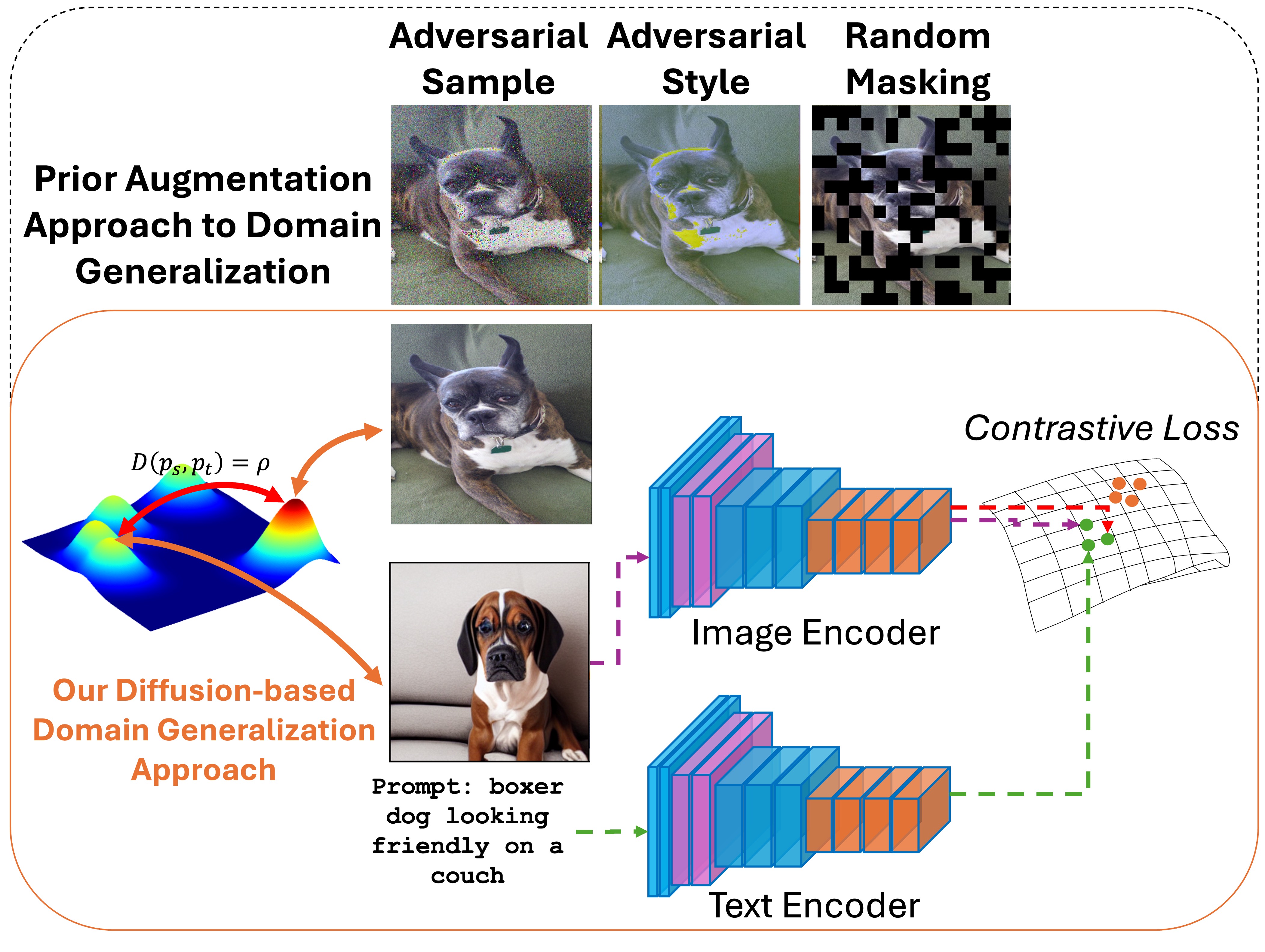}
    \vspace{-6mm}
    \caption{Comparison between Our Proposed Diffusion-based Domain Generalization with Prior Methods \cite{volpi2018generalizing, zhong2022adversarial, li2023scaling}.}\label{fig:abstract}
\end{wrapfigure}
The vision-language foundation models trained based on contrastive learning and exemplified by CLIP \cite{radford2021learning}, have gained more attention due to their outstanding performance on various tasks.
Although the vision-language foundation models have shown advantages on various downstream visual tasks, limited studies investigate their generalizability. 
Meanwhile, the generalizability of the foundation models still majorly relies on the large-scale pre-training datasets.
While many prior studies \cite{guo2023domaindrop, hu2023dandelionnet, yao2022pcl, li2023simple, long2023rethinking, zhang2023adanpc, chang2023domain, wang2023sharpness, chen2023metacausal} have been introduced to domain generalization for classification \cite{lee2023decompose, yao2022pcl, volpi2018generalizing, bui2021exploiting, li2023intra}, detection \cite{vidit2023clip, lin2021domain}, semantic segmentation \cite{zhong2022adversarial, lee2022wildnet, ding2023hgformer, huang2023style}, there are limited studies that address the domain generalization problem in the vision-language foundation model.
Despite being trained on a large-scale dataset, the generalizability of the vision-language foundation model has to be considered because it is a key factor in guaranteeing the performance of models against unknown data distributions.
The domain generalization approaches are urgently needed for foundation model training to ensure optimal performance and generalizability.
The current vision-language foundation models trained using contrastive learning often rely on data augmentations to improve their robustness and prevent overfitting.
However, these methods are not effective enough to improve the generalization of the foundation model. 
In particular, to improve the performance of CLIP models, most of the prior visual foundation models perform the data augmentation on visual inputs \cite{li2023scaling, mu2022slip, radford2021learning, nguyen2023micron, jia2021scaling} to increase the number of training samples and create challenging samples. 
These augmentation methods aim to increase the diversity of the data, thus enhancing the generalization of the foundation models. 
However, these visual augmentations concentrate on pixel-level modification like masking, adversarial perturbations, adversarial styles, or color jittering, which have a limited impact on enriching the semantic information of visual concepts.
Therefore, the generalizability to unknown data distributions of vision-language models remains limited.

In recent years, in parallel with the development of vision-language models, the diffusion model has shown its outstanding performance in data distribution modeling and generative AI. The diffusion approach, designed based on the nonequilibrium thermodynamics \cite{ho2020denoising}, is able to model the data distribution via the parameterized Markov chain trained using variational inference. Hence, the diffusion models can synthesize novel, high-quality, and complex data.
Moreover, the diffusion models are also able to efficiently model the conditional data distributions, e.g., text-to-image diffusion \cite{rombach2022high}.
Inspired by the success of diffusion, this paper fundamentally investigates its role and relation to the generalizability of the vision-language foundation models. In particular, we first model the domain generalization of the vision-language foundation model via the worst-case formula over data distributions that are near the training domain of the latent space. Then, using the Lagrangian relaxation and diffusion properties, we introduce a novel transformation approach that improves the generalizability of the vision-language foundation model by expanding training data distribution via the diffusion model.
Our theoretical analysis has shown the proposed approach is robust and well-generalized. It also has a better domain generalization compared to prior methods \cite{volpi2018generalizing, zhong2022adversarial}.

\textbf{Contributions:}
This paper introduces a novel \textbf{\textit{Diffusion-based Domain Generalization}} approach, a simple yet effective approach to improving the generalizability of the vision-language model, i.e., CLIP, by exploiting the power of the diffusion model (Fig. \ref{fig:abstract}). 
In particular, first, we form the domain generalization problem of the vision-language model via the worst-case formula over the training data distribution. By modeling the data conditional distribution via the diffusion model, we further provide a complete theoretical analysis of the relation of the diffusion model to adversarial augmentation.
Second, we introduce a new simple yet efficient \textbf{\textit{Transport Transformation}} to diffusion sampling that can synthesize adversarial samples to improve the generalizability of the vision-language model.
Thanks to our proposed Transport Transformation, our approach efficiently expands the training data distributions, therefore improving the ability to generalize to unseen data distributions of the vision-language model.
Finally, our extensive experiments on pre-training vision-language datasets at different scales, including CC3M, CC12M, and LAION400M, have shown the robustness of the proposed approach. Our approach has improved the performance of CLIP significantly on various benchmarks and outperformed other augmentation and domain generalization approaches. The theoretical analysis and empirical results guarantee that the proposed approach is simple yet scalable and contributes to the generalizability improvement of vision-language foundation models.

\vspace{-2mm}
\section{Related Work}
\vspace{-2mm}

\textbf{Vision-Language Foundation Model} 
The contrastive language-image training \cite{radford2021learning, jia2021scaling, yu2022coca, luo2023lexlip,wang2023equivariant, dehdashtian2024fairerclip} has become a prominent approach in developing the large-scale vision-language model \cite{radford2021learning, jia2021scaling}. 
CLIP \cite{radford2021learning} and ALIGN \cite{jia2021scaling}  first introduced contrastive learning to learn strong representations of images and texts for cross-modal alignment.
CoCa \cite{yu2022coca} proposed an additional decoder and generative image captioning.
SLIP \cite{mu2022slip}, DeCLIP \cite{li2021supervision}, FLIP \cite{li2023scaling} further improve the performance by using self-supervised training techniques.
LaCLIP \cite{fan2023improving} improved the performance of CLIP by introducing text augmentation via the large language model.
LiT \cite{zhai2022lit} and BASIC \cite{pham2023combined} improve the zero-shot transfer ability via further fine-tuning the language encoder.
SimVLM \cite{wang2021simvlm}, OFA \cite{wang2022ofa}, and BLIP \cite{li2022blip} train the vision-language model within an encoder-decoder framework with language generative losses. 
SigLIP \cite{zhai2023sigmoid} proposed a Sigmoid loss to compute the image-text similarity.

\textbf{Denoising Diffusion Probabilistic Model} (DDPM) has achieved state-of-the-art performance in density estimation and image synthesis \cite{ho2020denoising, rombach2022high}. 
The DDPM model defines a Markov chain of diffusion steps to gradually add random noise to data followed by learning to reverse the diffusion process via the UNet \cite{ho2020denoising} to construct the data sample from noise.
Subsequent studies further improved by reweighing the learning objective \cite{kim2024training}, improving the variance schedule \cite{nichol2021improved}, using distillation \cite{meng2023distillation}.
Denoising diffusion implicit models (DDIM) \cite{song2021denoising} was introduced to accelerate the sampling process by generalizing DDPMs.
Meanwhile, the Latent Diffusion Model (LDM) \cite{rombach2022high} proposed a two-stage diffusion model where the diffusion process is performed on the latent space.
Other approaches improve the DDPMs by introducing cascaded generation \cite{ho2022cascaded}, incorporating with GANs \cite{wang2023diffusiongan}, using wavelet transformation \cite{phung2023wavelet}, and introducing momentum-based diffusion \cite{dockhorn2021score, huang2022prodiff}.
The diffusion model also has 
an ability of conditional synthesis, e.g., text-to-image \cite{rombach2022high, ramesh2022hierarchical, saharia2022photorealistic}, image editing \cite{kawar2023imagic, nguyen2023visual}. This conditional ability can be implemented as explicit conditions \cite{rombach2022high}, classifier guidance \cite{nie2021controllable, dockhorn2021score, dhariwal2021diffusion}, or classifier-free guidance \cite{ho2022classifier}.
The later studies further improve diffusion models by introducing a single-step diffusion \cite{hoang2023swiftbrush}, subject-driven fine-tuning \cite{ruiz2023dreambooth}.

\textbf{Domain Generalization} 
aims to learn a robust model from single or multiple source data so that the model can later be well generalized to unseen data domains.
One stream of the domain generalization approach focuses on using data augmentation to improve the generalizability of the model \cite{li2023scaling, hoyer2023domain}.
Recent studies adopted image masking \cite{li2023scaling} or the image-editing technique or style transfer via the diffusion model to improve the performance of object classification \cite{trabucco2023effective, fu2024dreamda, feng2023diverse, diffuseMix2024}, object detection \cite{fang2024data}, or 3D classification \cite{shen2024diffclip}.
Another stream of domain generalization focuses on learning the invariant feature space by jointly optimizing a multi-domain autoencoder \cite{domain_generalization_Ghifary_2015_ICCV, domain_generalization_Li_2018_CVPR}, removing domain specific via normalization \cite{ulyanov2016instance, fan2021adversarially, tang2020selfnorm}, learning in the frequency domain \cite{wang2022domain, huang2021fsdr, lin2023deep}.
Adversarial training has been introduced to learn the robust model by forming novel domains via the generated adversarial samples.
Adversarial Data Augmentation (ADA) \cite{volpi2018generalizing} first introduced an approach to generate adversarial samples via max-min iterative training. 
Later, M-ADA \cite{qiao2020learning} further improved ADA by training an additional autoencoder.
Other approaches learn the domain-invariant features with adversarial samples via meta-learning \cite{finn2017model, qiao2021uncertainty}, or image-style adversarial learning \cite{zhong2022adversarial}.
In addition, another domain generalization approach improves the generalization ability by re-designing the deep neural network \cite{li2021progressive, pan2018two} or using an ensemble of expert models \cite{arpit2022ensemble}.
To the best of our knowledge, \textit{these prior studies have not fully investigated the fundamentals of diffusion to domain generalization of the vision-language foundation models. Therefore, in this paper, we provide a theoretical analysis of diffusion to the generalizability of the vision-foundation model, followed by proposing a new simple yet efficient diffusion-based domain generalization approach.}

\section{Theoretical Analysis of Generalizability in Foundation Model}

\subsection{Preliminary}
\label{sec:diffusion_background}

\textbf{Diffusion Model} 
formulates the data distribution $p(\mathbf{x})$ by gradually denoising a normally distributed variable via the reverse process of a fixed Markov Chain of length $T$, i.e., $p(\mathbf{x}_0) = \int p(\mathbf{x}_{0:T})d\mathbf{x}_{1:T}$,  with a Gaussian transition starting at $p(\mathbf{x}_T) = p(\mathbf{z}) = \mathcal{N}(\mathbf{z}; \mathbf{0}, \mathbf{I})$
The diffusion model includes the forward and backward processes. The forward diffusion process, i.e., $q(\mathbf{x}_i|\mathbf{x}_{i-1})$ is defined as:
\begin{equation}
\small
    q(\mathbf{x}_{1:T}|\mathbf{x}_0) = \prod_{i=1}^{T} q(\mathbf{x}_i |\mathbf{x}_{i-1}) \quad q(\mathbf{x}_{i} | \mathbf{x}_{i-1}) = \mathcal{N}(\mathbf{x}_i, \sqrt{1-\beta_i}\mathbf{x}_{t-1}, \beta_i\mathbf{I})
\end{equation}
where $\beta_i$ is a variance schedule. Then, the backward process, i.e., $p(x_{k-1}|x_{k-1})$, is defined as:
\begin{equation} \label{eqn:diffusion_backward}
\small
    p(\mathbf{x}_{0:T}) = p(\mathbf{x}_T) \prod_{i=1}^{T} p(\mathbf{x}_{i-1}|\mathbf{x}_i) \quad p(\mathbf{x}_{i-1}|\mathbf{x}_i) = \mathcal{N}(\mathbf{x}_{i-1}; \boldsymbol{\mu}_{\theta}(\mathbf{x}_i, i), \boldsymbol{\Sigma}_{\theta}(\mathbf{x}_i, i)) 
\end{equation}
The backward process adopts
a denoising model $\epsilon_{\theta}$ to predict the denoised variant from $\mathbf{x}_i$.
Then, 
the model is learned via 
the usual variational bound on negative log-likelihood is as follows:
\begin{equation}
\small
    \theta^* = \arg\min_{\theta} \mathbb{E}_{\mathbf{x}, \epsilon \in \mathcal{N}(\mathbf{0, I}), i} \left[|| \epsilon - \epsilon_{\theta}(\mathbf{x}_i, i) ||_2^2 \right]
\end{equation}
where $\theta$ is the parameter of $\epsilon$, $\mathbf{x}_i = \sqrt{\overline{\alpha}_i}\mathbf{x} + \sqrt{1 - \overline{\alpha}_i}\epsilon$, $\alpha_i = 1 -\beta_i$, $\overline{\alpha}_i = \prod^i_{s=1}\alpha_s$, and $i$ is uniformly sampled from $1$ to $T$, i.e., $i \in \mathcal{U}(1, T)$.
The diffusion model is capable of modeling the conditional distribution, i.e., $p(\mathbf{x}|\mathbf{p})$ where $\mathbf{p}$ is the condition (e.g., a text prompt).
This ability can be done by implementing a conditional denoising model $\epsilon_{\theta}(\mathbf{x}_i, i, \mathbf{p})$.

\textbf{Contrastive Language-Image Pretraining (CLIP)} \cite{radford2021learning} has shown its outstanding performance in the training vision-language foundation model using language supervision. Formally, let $\mathbf{x}, \mathbf{p} \sim p(\mathbf{x}, \mathbf{p})$ be the source training data of the CLIP model where $\mathbf{x}$ is the image, and $\mathbf{p}$ is the corresponding prompt,  $F^{\mathbf{x}}$ and $F^{\mathbf{p}}$ be the vision and language encoder, and $\mathbf{f}^{\mathbf{x}}$ and $\mathbf{f}^{\mathbf{p}}$ be the features extracted by the vision and language encoder, respectively, i.e., $\mathbf{f}^{\mathbf{x}} = F^{\mathbf{x}}(\mathbf{x})$ and $\mathbf{f}^{\mathbf{p}} = F^{\mathbf{p}}(\mathbf{p})$. The CLIP model is learned via contrastive loss, where the pairs of images and corresponding texts are the positive pairs. The CLIP model can formulated as follows:
\begin{equation}\label{eqn:clip_model_learning}
\small
\theta^*_{F^{\mathbf{x}}}, \theta^*_{F^{\mathbf{p}}} = \arg\min_{\theta_{F^{\mathbf{x}}}, \theta_{F^{\mathbf{p}}}} \mathbb{E}_{\mathbf{x}, \mathbf{p}_t \sim p(\mathbf{x}, \mathbf{p})} -\log \frac{\exp({\operatorname{sim}(F^{\mathbf{x}}(\mathbf{x}), F^{\mathbf{p}}(\mathbf{p}))}/{\tau})}{\sum_{k}\exp({\operatorname{sim}(F^{\mathbf{x}}(\mathbf{x}), F^{\mathbf{p}}(\mathbf{p}^k))}/{\tau})}
\end{equation}
where $\theta_{F^{\mathbf{x}}}, \theta_{F^{\mathbf{p}}}$ are parameters of $F^{\mathbf{x}}$ and $F^{\mathbf{p}}$, $\mathbf{p}^k$ is the negative text sample of $\mathbf{x}$, $\tau$ is the a temperature to scale logits, $\operatorname{sim}$ is the dot product to measure distance between features. For simplicity, Eqn. \eqref{eqn:clip_model_learning} only illustrates the contrastive loss over images. In practice, a symmetrical loss over texts is also applied, and the loss is the average of the contrastive loss over images and texts.

\subsection{Domain Generalization of Contrastive Language-Image Pre-Training}

In our paper, we aim to develop a domain generalization approach to CLIP that is able to better generalize to new unknown data distributions. In this work, we consider the training data of CLIP drawn from a single source data \cite{volpi2018generalizing}, i.e., $\mathbf{x}_s, \mathbf{p}_s \in p(\mathbf{x}_s, \mathbf{p}_s)$. Inspired by prior work in robust optimization, we propose to model the domain generalization of CLIP via the worst-case problem around the source data distribution $p(\mathbf{x}_s, \mathbf{p}_s)$ as follows:
\begin{equation} \label{eqn:dg_general}
\small
    \theta^*_{F^{\mathbf{x}}}, \theta^*_{F^{\mathbf{p}}} = \arg\min_{\theta_{F^{\mathbf{x}}}, \theta_{F^{\mathbf{p}}}} \sup_{p_t:\mathcal{D}(p_t, p_s)\leq \rho} \mathbb{E}_{\mathbf{x}_t, \mathbf{p}_t \sim p_t(\mathbf{x}_t, \mathbf{p}_t)} -\log \frac{\exp({\operatorname{sim}(F^{\mathbf{x}}(\mathbf{x}_t), F^{\mathbf{p}}(\mathbf{p}_t))}/{\tau})}{\sum_{k}\exp({\operatorname{sim}(F^{\mathbf{x}}(\mathbf{x}_t), F^{\mathbf{p}}(\mathbf{p}^k))}/{\tau})}
\end{equation}
where $\mathbf{x}_t, \mathbf{p}_t$ are images and prompt sampled from $p_t$, 
$\mathcal{D}(p_s, p_t)$ is the Wasserstein metric measure the distance between two data distributions $p_s$ and $p_t$, $\rho$ is the distance constraint, $p_t$ is $\rho$-away unknown data distributions from $p_s$, i.e., $\mathcal{D}(p_s, p_t) \leq \rho$.
Eqn. \eqref{eqn:dg_general} aims to guarantee good performance of the CLIP model against the unknown data distribution.

\textbf{Domain Generalization of CLIP} In our paper, we are interested in the problem of domain generalization of CLIP where we aim to improve the performance of CLIP, especially when using the CLIP model for downstream tasks, e.g., zero-shot classification, linear probing, or fine-tuning. In this learning scenario, since the target data distribution $p_t$ is completely unknown, the hyper-parameter $\rho$ plays an important role since it will indicate the generalizability of the CLIP model to new data (or test domains). 
To solve the Eqn \eqref{eqn:dg_general}, the Lagrange multiplier can be adopted to reform Eqn \eqref{eqn:dg_general} as:
\begin{equation}\label{eqn:adv_sample}
\small
    \mathbf{x}^*_t =\arg \max_{\mathbf{x}_t} \Big\{ \mathcal{L}_{CLIP}(\mathbf{x}_t,\mathbf{p}_s) - \lambda \mathcal{D}(p_t(\mathbf{x}_t, \mathbf{p}_s), p_s(\mathbf{x}_s, \mathbf{p}_s)) \Big\}
\end{equation}
where $\mathbf{x}^*_t$ is the \textbf{\textit{adversarial sample}} (corresponding to prompt $\mathbf{p}_s$) to improve the generalization and robustness of the CLIP model,
$\mathcal{L}_{CLIP}$ is the contrastive language-image pretraining loss defined in Eqn. \eqref{eqn:clip_model_learning}, 
$\lambda$ is the hyper-parameter that is inverse proportional of $\rho$, $\mathcal{D}(p_t(\mathbf{x}_t, \mathbf{p}_s), p_s(\mathbf{x}_s, \mathbf{p}_s))$ is the transportation cost that moving from the $\mathbf{x}_s, \mathbf{p}_s \sim p_s(\mathbf{x}_s, \mathbf{p}_s)$ to the distribution $p_t$.
Since our paper aims to improve the generalizability of the CLIP model on the downstream vision tasks, the scope of this work focuses on the adversarial sample in the vision domain. 
Eqn. \eqref{eqn:adv_sample} aims to create augmented samples so that the distribution of augmented samples is $\rho$-away from the original one and increases contrastive learning loss. Then, using these augmented samples will potentially improve the generalizability of CLIP. 

\begin{figure}[!t]
    \centering
    \includegraphics[width=0.9\textwidth]{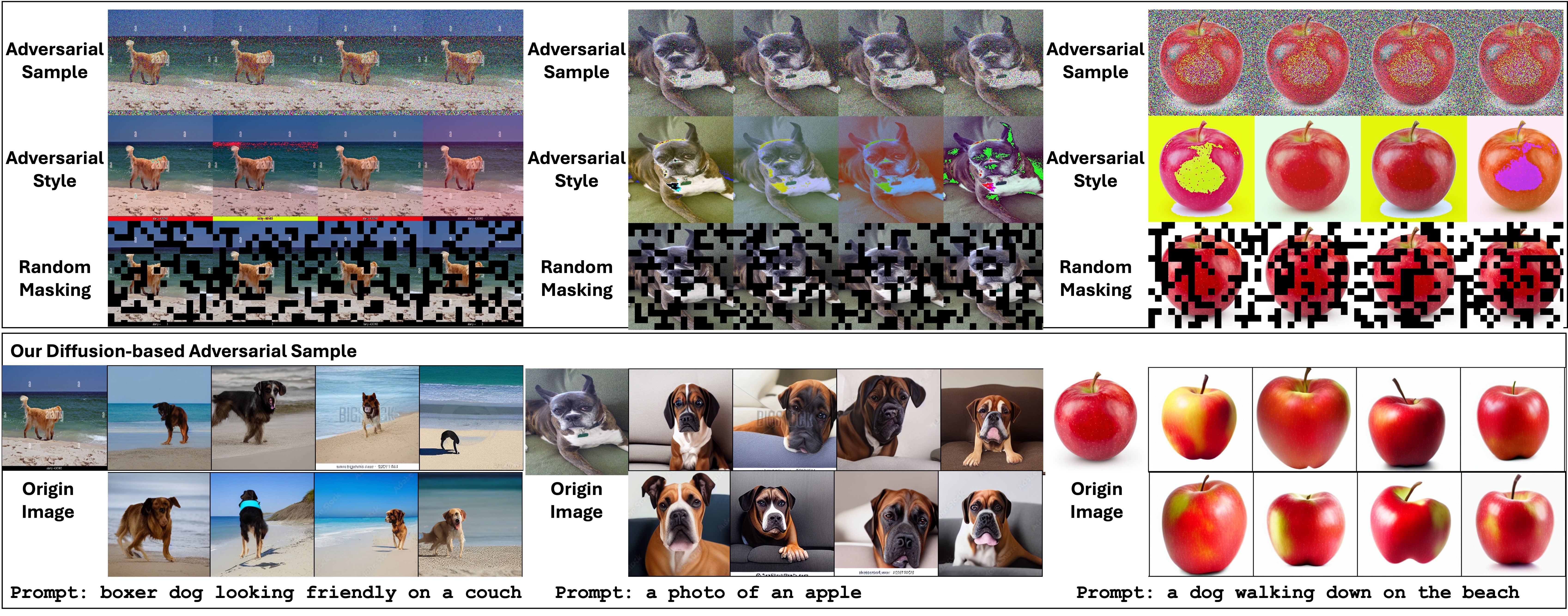}
    \vspace{-2mm}
    \caption{The Comparison of Our Diffusion-based Adversarial Sample and Prior Augmentations (Adversarial Sample \cite{volpi2018generalizing}, Adversarial Style \cite{zhong2022adversarial}, Masking Sample \cite{li2023scaling}).}
    \label{fig:adv_sample_compare}
    \vspace{-6mm}
\end{figure}

\textbf{Limitation of Prior Work}
Prior work adopts adversarial training \cite{volpi2018generalizing}, augmentation methods \cite{li2023scaling}, or adversarial style augmentation \cite{zhong2022adversarial} to generate adversarial/augmented samples to improve the generalizability.
Although prior results have shown the potential performance improvement,
these approaches remain limited in terms of expanding their generalizability to unknown distributions.
Indeed, adversarial learning \cite{volpi2018generalizing, zhong2022adversarial} tries to add the perturbation via maximizing loss or adversarial styles into images.
Meanwhile, the augmentation methods create different variations of images by performing heuristic pixel-wise image operations (e.g., masking, cropping, color jittering, etc).
However, the data distributions of augmented samples generated by prior methods \cite{volpi2018generalizing, zhong2022adversarial, li2023scaling} 
remain unchanged or little changed compared to the original data distribution.
This can be explained since, despite the different variations of augmented samples, the content information, e.g., object appearances, shapes, etc., and the semantic background information remained the same.
For example, as shown in Fig. \ref{fig:adv_sample_compare}, the target object of augmented samples created by \cite{volpi2018generalizing, zhong2022adversarial, li2023scaling} remains unchanged. The semantic background, in general, is similar to the original image, with noise added.

\subsection{The Relation of Diffusion to Adversarial Augmentation} 

As aforementioned, the goal of the adversarial sample in Eqn. \eqref{eqn:adv_sample} is to move the data sample from the source training $\mathbf{x}_s$ to the $\mathbf{x}^*_t$ in the $\rho$-away distribution so that maximize the contrastive language-image pretraining loss $\mathcal{L}_{CLIP}(\mathbf{x}_t, p_s)$. As shown in Eqn. \eqref{eqn:adv_sample}, the sample $\mathbf{x}^*_t$ is depending on the source training sample $\mathbf{x}_s$, the text prompt $\mathbf{p}_s$, and the distance between two distributions $\rho$. Therefore, 
in our work, we consider the adversarial sample $\mathbf{x}^*_t$ is draw from a $\rho$-away distribution conditioned on $\mathbf{x}_s$, $\mathbf{p}_s$, and $\rho$, i.e., $\mathbf{x}^*_t \in p(\mathbf{x}^*_t | \mathbf{x}_s, \mathbf{p}_s, \rho)$.

\begin{wrapfigure}[10]{r}{0.25\textwidth}
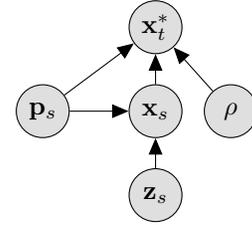

    \label{eqn:graphical_model}
    \centering
    \vspace{-8mm}
    \tikz{
         \node[obs] (xt) {$\mathbf{x}^*_t$};
         \node[obs,below=of xt,xshift=-0cm, yshift=0.6cm] (xs) {$\mathbf{x}_s$}; 
         \node[obs,below=of xs,xshift=-0cm, yshift=0.6cm] (zs) {$\mathbf{z}_s$}; 
         \node[obs,below=of xt,xshift=-1.5cm, yshift=0.6cm] (ps) {$\mathbf{p}_s$}; 
         \node[obs,below=of xt,xshift=1cm, yshift=0.6cm] (rho) {$\rho$}; 
         \edge {xs,ps,rho} {xt}
         \edge {zs,ps} {xs}
    }
    \vspace{-1mm}
    \caption{The Relation Between Adversarial Sample and Source Data.}
    \label{fig:graphic_model}
\end{wrapfigure}
\textbf{The Source Data Distribution} Since the image $\mathbf{x}_s$ and $\mathbf{p}_s$ is a pair of image and text, without a strict argument, we could assume that the image $\mathbf{x}_s$ is conditioned on the text prompt $\mathbf{p}_s$, i.e., $\mathbf{x}_s \in p(\mathbf{x}_s | \mathbf{p}_s)$. 
As presented in Sec. \ref{sec:diffusion_background}, the conditional distribution $p(\mathbf{x}_s | \mathbf{p}_s)$ could be efficiently modeled by the diffusion. Let $\mathbf{z}_s \in \mathcal{N}({\mathbf{0}, \mathbf{I}})$ be the latent variable of image $\mathbf{x}_s$. 
Then, the image $\mathbf{x}_s$ can be modeled via the backward process of diffusion conditioned on $\mathbf{p}_s$ and $\mathbf{z}_s$ as in Eqn. \eqref{eqn:diffusion_backward}
For simplicity, we rewrite the data distribution $\mathbf{x}_s \sim p(\mathbf{x}_{0:T} | \mathbf{p}_s)$ via the latent variable $\mathbf{z}_s$ as $\mathbf{x}_s \sim p(\mathbf{x}_s | \mathbf{z}_s, \mathbf{p}_s)$. 

\textbf{The Diffusion-based Adversarial Augmentation}
Fig. \ref{fig:graphic_model} illustrates the graphical model that define the relation among $\mathbf{x}_s$, $\mathbf{p}_s$, $\mathbf{z}_s$, $\mathbf{x}^*_t$, and $\rho$. The relation in this graphical model is established based on two conditions of the adversarial sample $p(\mathbf{x}^*_t | \mathbf{x}_s, \mathbf{p}_s, \rho)$ and the conditional diffusion model $p(\mathbf{x}_s | \mathbf{z}_s, \mathbf{p}_s)$.
As shown in the graphical model, we have observed that the adversarial sample $\mathbf{x}^*_t$ depends on $(\mathbf{x}_s, \mathbf{p}_s, \rho)$ while the image $\mathbf{x}_s$ is conditioned on $(\mathbf{z}_s, \mathbf{p}_s)$. Therefore, for simplicity, without a strict argument, we assume that the adversarial sample $\mathbf{x}^*_t$ is equivalently depending on $(\mathbf{z}_s, \mathbf{p}_s, \rho)$ defined as follows
\begin{equation}
\begin{split}
\small
    \mathbf{x}^*_t &\sim p(\mathbf{x}^*_t | \mathbf{x}_s, \mathbf{p}_s, \rho) \Rightarrow \mathbf{x}^*_t \sim p(\mathbf{x}^*_t | \mathbf{z}_s, \mathbf{p}_s, \rho) \Rightarrow \mathbf{x}^*_t \sim p(\mathbf{x}^*_t | \mathbf{z}^*_t, \mathbf{p}_s) \; \text{where} \; \mathbf{z}^*_t = \mathcal{T}(\mathbf{z}_s, \rho) \\
\end{split}
\end{equation}
where $\mathcal{T}$ is the transport transformation on the latent space.
Intuitively, instead of moving the image $\mathbf{x}_s$ in the image space to $\mathbf{x}^*_t$ in the new distribution with a transportation cost of $\mathcal{D}(p_t(\mathbf{x}^*_t, \mathbf{p}_s), p_s(\mathbf{x}_s, \mathbf{p}_s))$ as in Eqn. \eqref{eqn:adv_sample} which is a challenging problem,
we are going to move the latent variable $\mathbf{z}_s$ to $\mathbf{z}^*_t$ via the transport function $\mathcal{T}$ controlled by $\rho$.
Since the latent space of the diffusion model is tractable (as it is a Gaussian distribution), moving $\mathbf{z}_s$ to $\mathbf{z}^*_t$ on latent space is controllable and easier than moving samples on the image space.
Then, the adversarial sample of $\mathbf{x}^*_t$ can be achieved by the reverse process of the diffusion model.
With our proposed diffusion-based augmentation approach, thanks to the power of the diffusion model \cite{rombach2022high}, our approach is able to synthesize novel adversarial samples that still maintain the semantic conditions on the prompt $\mathbf{p}_s$ while being effectively used to improve the generalizability in training CLIP model.
As shown in Fig. \ref{fig:adv_sample_compare}, our proposed approach can generalize a new sample with the sample condition prompt, but the content and semantic background of the image have been changed significantly. This helps to strongly expand the data distribution during training to improve the generalizability of unknown data distribution.

\vspace{-2mm}
\subsection{The Proposed Transport Transformation} 

It is important to design a transformation $\mathcal{T}$ that satisfies the condition of domain generalization, i.e., $\mathcal{D}(p_s(\mathbf{x}_s, \mathbf{p}_s), p_t(\mathbf{x}_t, \mathbf{p}_t)) \leq \rho$ in Eqn. \eqref{eqn:dg_general}, to guarantee the generalizability defined in Eqn. \eqref{eqn:dg_general}.
Since the data distribution in our approach is displaced in the latent space of $\mathbf{z}_s$, with a strict argument, the condition of domain generalization via the latent space could be written as follows:
\begin{equation}\label{eqn:requirement}
    \small
    \mathcal{D}(p_s(\mathbf{x}_s, \mathbf{p}_s), p_t(\mathbf{x}^*_t, \mathbf{p}_t)) \propto \mathcal{D}(p_s(\mathbf{z}_s), p_t(\mathbf{z}^*_t)) \leq \rho
\end{equation}

In our proposed approach, in order the meet the requirement as defined in Eqn. \eqref{eqn:requirement}, the transport transformation $\mathcal{T}$ can be defined as follows:
\begin{equation} \label{eqn:transform_function}
    \small
    \mathbf{z}^*_t = \mathcal{T}(\mathbf{z}_s, \rho) = \frac{\mathbf{z}_s + \mathcal{N}(\alpha\sqrt{2}, \mathbf{I})}{\sqrt{2}} \quad \text{where } \alpha \sim \mathcal{U}(-\rho, \rho) 
\end{equation}
where $\alpha$ is controllable hyper-parameter uniformly sampled from $\mathcal{U}(-\rho, \rho)$.

\textbf{\textit{Proposition 1:}} \textit{Given $\mathbf{z}_s \in \mathcal{N}(\mathbf{0, I})$ and $\alpha$ ($-\rho \leq \alpha \leq \rho$), the condition of distance between distributions $\mathcal{D}(p_s(\mathbf{z}_s), p_t(\mathbf{z}^*_t)) \leq \rho$ holds if the transport transformation $\mathcal{T}$ is defined as $\mathbf{z}^*_t = \mathcal{T}(\mathbf{z}_s, \rho) = \frac{\mathbf{z}_s + \mathcal{N}(\alpha\sqrt{2}, \mathbf{I})}{\sqrt{2}}$. The proof is provided in the appendix.}

While there could be multiple transport transformations that satisfy the condition of the distance between two distributions, i.e.,  $\mathcal{D}(p_s, p_t) \leq \rho$, we have observed that our proposed metric in Eqn. \eqref{eqn:transform_function} provides a better mechanism to move the sample on the latent spaces. This could be explained since our metric is able to expand the training data distribution by moving the original latent vectors in the latent space while still maintaining the important property as mentioned in \textbf{Proposition 1}. In addition, by moving the latent vector $\mathbf{z}_s$ in the latent space with a controlled parameter $\rho$, our metric can guarantee the semantic content information compared to the original one while creating the diverse semantic variations of the images. This also encourages the diffusion model to avoid synthesizing useless random images with uncontrolled latent vectors.

\begin{figure}[!t]
    \centering
    \includegraphics[width=0.9\textwidth]{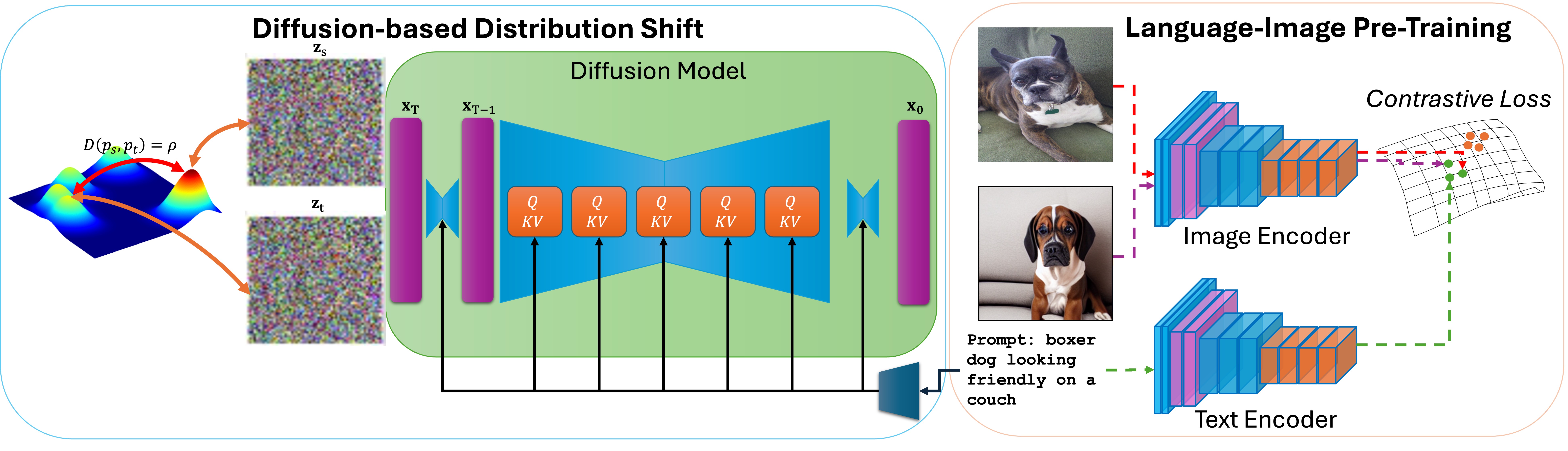}
    \vspace{-2mm}
    \caption{The Proposed Diffusion-based Domain Generalization Framework}
    \label{fig:da_framework}
    \vspace{-6mm}
\end{figure}

\section{The Proposed Diffusion-based Domain Generalization Training Approach}

\textbf{Large Scale Diffusion-based Augmentation Sample Generation}
As shown in our theoretical analysis, generating our diffusion-based adversarial samples does not require alternative training steps with the CLIP training procedure. We have empirically observed that retraining the text-to-image diffusion model is unnecessary because the pre-trained diffusion model has been well learned on extreme-scale datasets, can model the data distribution well, and generates diverse synthetic data. Therefore, in our approach, we adopt the pretrained Latent Diffusion model \cite{rombach2022high} to generate the adversarial samples in advance to save the training time of CLIP. Formally, for each image $\mathbf{x}_s$ and its corresponding prompt $\mathbf{p}_s$, we are going to generate $M$ different augmented samples $\mathbf{x}^*_t$ via the latent diffusion model by the following process:
\begin{equation}
\small
\begin{split}
    \mathbf{z}_s &= \operatorname{LDMForward}(\mathbf{x}_s); \;\;
    \mathbf{z}^*_t = \frac{\mathbf{z}_s + \mathcal{N}(\alpha\sqrt{2}, \mathbf{I})}{\sqrt{2}} \text{ where } \alpha \sim \mathcal{U}(-\rho, \rho);  \;\;
    \mathbf{x}^*_t = \operatorname{LDMBackward}(\mathbf{z}_t)
\end{split}
\end{equation}
where $\operatorname{LDMForward}$ and $\operatorname{LDMBackward}$ are the forward and backward processes of the latent diffusion model. Generating the adversarial samples during training will result in a longer training time for CLIP, which is unnecessary. Therefore, we propose to generate the adversarial samples via diffusion in advance, followed by using them to train the CLIP model, which is more time-efficient.

\textbf{The Diffusion-based Domain Generalization Training}
Fig. \ref{fig:da_framework} illustrates our proposed domain generalization framework. After the generation steps, each real image has $M$ different adversarial samples. Then, we are able to improve the generability of the CLIP model by training on the real and adversarial samples together. Formally, learning the CLIP model can be re-written as follows: 
\begin{equation} \label{eqn:domain_adaptation}
\small
        \theta^*_{F^{\mathbf{x}}}, \theta^*_{F^{\mathbf{p}}} = \arg\min_{\theta_{F^{\mathbf{x}}}, \theta_{F^{\mathbf{p}}}} \mathbb{E}_{\mathbf{x}_s, \mathbf{p}_s, \mathbf{x}^*_t} \left[\mathcal{L}_{CLIP}(\mathbf{x}_s, \mathbf{p}_s) + \mathcal{L}_{CLIP}(\mathbf{x}^*_t, \mathbf{p}_s) \right]
\end{equation}

\section{Experiments}

\subsection{Datasets, Implementations, and Evaluations}

\textbf{Dataset}
We trained our foundation model on three different image-text datasets at different scales: Conceptual Captions 3M (CC3M)  \cite{sharma2018conceptual}, Conceptual Captions 12M (CC12M) \cite{changpinyo2021cc12m}, and LAION400M \cite{schuhmann2021laion}.
Due to the hardware constraints, our ablation studies are mainly conducted on CC3M and CC12M. 
We evaluate our models on ImgageNet 1K \cite{imagenet15russakovsky} and six common datasets, including, STL-10 \cite{coates2011analysis}, Country-211 \cite{thomee2016yfcc100m},	Caltech-101 \cite{fei2006one}, Flowers \cite{nilsback2006visual}, Pets \cite{parkhi2012cats}, and SUN-397 \cite{xiao2016sun}.

\textbf{Implementation}
We adopt the implementation of OpenCLIP \cite{cherti2023reproducible} and Latent Diffusion \cite{rombach2022high} in our experiments. 
For the CLIP model, we use the ViT-B/16 architecture.
The results of other network backbones are reported in the appendix. For a fair comparison, our model is trained for 32 epochs with a similar hyper-parameter setting as \cite{radford2021learning, cherti2023reproducible}. 
We utilize $32$ NVIDIA A100 GPUs (40GB), and the batch size of our experiments is set to $320$ per GPU. 
For image synthesis, we use the text-to-image latent diffusion model \cite{rombach2022high} to generate images at the resolution of $256 \times 256$ with $10$ DDIM steps.
For each real image, we generate $M=10$ different synthetic images.
The controlling hyper-parameter of the distance between distributions $\rho$ is set to $0.5$ in our experiments.
Due to time and hardware constraints, we choose to use only $10$ DDIM generation steps. This offers image quality that meets acceptable standards \cite{rombach2022high} while maintaining efficient data generation time on large-scale datasets (e.g., approximately 7.5 hours to generate 12M adversarial samples of CC12M on 32 GPUs).

\begin{wraptable}[11]{l}{0.4\textwidth}
\centering
\vspace{-6mm}
\caption{The Effectiveness of Distribution Moving $\rho$.}
\label{tab:rho_ablation}
\vspace{-2mm}
\resizebox{0.4\textwidth}{!}{
    \begin{tabular}{l|c|ccc}
    \hline
                           & $\rho$ & Zeroshot & Linear Prob & Fine-Tune \\
    \hline
    \multirow{5}{*}{\rotatebox{90}{CC3M}}  & 0.05          & 17.28    & 54.13       & 80.08     \\
                           & 0.20          & 18.79    & 55.11       & 80.61     \\
                           & 0.50          & \textbf{20.33}    & \textbf{56.14}       & \textbf{81.12}     \\
                           & 0.70          & 19.82    & 55.44       & 80.09     \\
                           & 1.00          & 16.68    & 52.92       & 79.14     \\
    \hline
    \multirow{5}{*}{\rotatebox{90}{CC12M}} & 0.05          & 36.44    & 69.27       & 80.28     \\
                           & 0.20          & 38.37    & 71.17       & 83.11     \\
                           & 0.50          & \textbf{39.34}    & \textbf{72.12}       & \textbf{84.67}     \\
                           & 0.70          & 37.34    & 69.12       & 82.89     \\
                           & 1.00          & 35.19    & 68.94       & 81.74    \\
    \hline
    \end{tabular}
}
\end{wraptable}
\textbf{Evaluation Setup}
In our experiments, we consider three different evaluation metrics, i.e., Zero-shot Classification Accuracy, Linear Probing Accuracy, and Fine-tuning Accuracy. 
For zero-shot classification, we adopt the template of prompts and evaluation protocol as described in CLIP \cite{radford2021learning}. 
For linear probing, following the common practices \cite{radford2021learning, li2023scaling, mu2022slip}, we use our frozen pre-trained image encoder to extract features followed by training a linear classifier. 
For a fair comparison, we adopt the hyper-parameter setting from \cite{radford2021learning, cherti2023reproducible}. 
For fine-tuning evaluation, we fine-tune the end-to-end image encoder with a linear classifier on the ImageNet 1K dataset. We adopt the implementation and learning hyper-parameter setting from \cite{cherti2023reproducible} for fair comparisons.
The majority of our experiments are evaluated on the ImageNet 1K dataset.
To further illustrate the generability of our model, we also perform the zero-shot evaluation on six different zero-shot benchmarks STL-10, Country-211, Caltech-101, Flowers, Pets, and SUN-397.

\subsection{Ablation Studies}

\begin{wraptable}[10]{r}{0.4\textwidth}
\centering
\vspace{-15mm}
\caption{The Effectiveness of Number of Generated Samples.}
\label{tab:number_of_image_abl}
\vspace{-2mm}
\resizebox{0.4\textwidth}{!}{
    \begin{tabular}{l|c|ccc}
    \hline
                           & {$M$} & {Zeroshot} & {Linear Prob} & {Fine-Tune} \\
    \hline
    \multirow{6}{*}{\rotatebox{90}{CC3M}}  & 
    0                   & 17.10          & 53.50          & 79.50  \\
    
    & 3                     & 18.36                        & 54.05                           & 79.90                         \\
                           & 5                     & 19.05                        & 55.17                           & 80.82                         \\
                           & 10                    & \textbf{20.33}               & \textbf{56.14}                  & \textbf{81.12}                \\
                           & 15                    & 20.40                        & 57.26                           & 81.17                         \\
                           & 20                    & 20.28                        & 56.18                           & 81.11                         \\
    \hline
    \multirow{6}{*}{\rotatebox{90}{CC12M}} & 0 & 36.50          & 69.00          & 82.10 \\
    & 3                     & 37.34                        & 70.25                           & 82.84                         \\
                           & 5                     & 38.10                        & 71.44                           & 83.58                         \\
                           & 10                    & \textbf{39.34}               & \textbf{72.12}                  & \textbf{84.67}                \\
                           & 15                    & 39.21                        & 72.18                           & 84.65                         \\
                           & 20                    & 39.49                        & 72.15                           & 84.68        \\
        \hline
    \end{tabular}
}
\end{wraptable}

\textbf{Effectiveness of Distribution Moving $\rho$}
The results in Table \ref{tab:rho_ablation} illustrate the effectiveness of the distance between distribution $\rho$. When the value of $\rho$ is small, i.e., $\rho = 0.05$, the CLIP gains a little improvement due to the small distribution shift. Then, the performance is gradually improved when the value of $\rho$ increases from $0.05$ to $0.5$.
When the value of $\rho$ is increased, the CLIP model can improve its generalizability to unknown distributions. 
However, if we keep increasing the value of $\rho$, the performance tends to drop. This is because if we shift the new data distribution in the latent space far away from the original data distribution ($\mathcal{N}(\mathbf{0, I})$), the quality of synthetic images generated by the latent diffusion model will dramatically drop in both realism and content information. Our best performance of CLIP is achieved at $\rho$ of $0.5$. 

\textbf{Effectiveness of Number of Augmented Images}
As shown in Table \ref{tab:number_of_image_abl}, the performance of our domain generalization approach evaluated on ImageNet1K is gradually increased when the number of adversarial images is increased.
When we use only $3$ adversarial images, the CLIP model gains a minor performance.
Meanwhile, when we use the $10$ adversarial images during training, the zero-shot classification performance of CLIP trained on CC3M and CC12M archives up to $20.33\%$ and $39.34\%$. The performance of linear probing and fine-tuning is also significantly improved when the number of adversarial images is increased.
However, if we keep increasing the number of images, we have observed that the performance of the CLIP model is becoming stable.
Therefore, generating $10$ adversarial images for each real image is a good trade-off between performance and time efficiency.

\begin{wraptable}[8]{l}{0.4\textwidth}
\centering
\setlength{\tabcolsep}{3pt}
\vspace{-5mm}
\caption{The Effectiveness of Transport Transformation.}
\label{tab:transport_abl}
\vspace{-2mm}
\resizebox{0.4\textwidth}{!}{
\begin{tabular}{c|l|ccc}
\hline
                       &                    & Zeroshot & Linear Prob & Fine-Tune \\
\hline
\multirow{3}{*}{\rotatebox{90}{CC3M}}  & CLIP & 17.10          & 53.50          & 79.50  \\
                      & Random              & 15.34    & 50.10       & 77.87     \\
                       & $\mathcal{T}$       & \textbf{20.33}    & \textbf{56.14}      & \textbf{81.12}     \\
\hline
\multirow{3}{*}{\rotatebox{90}{CC12M}} & CLIP & 36.50          & 69.00          & 82.10 \\
& Random              & 34.90    & 67.35       & 80.61     \\
                       & $\mathcal{T}$       & \textbf{39.34}    & \textbf{72.12}       & \textbf{84.67}    \\
\hline
\end{tabular}
}
\end{wraptable}

\noindent
\textbf{Effectiveness of Transport Transformation}
To illustrate the effectiveness of our Transport Transformation $T$, we 
compared it with another transformation. We define another random transformation by sampling $\mathbf{z}^*_t$ from the normal distribution $\mathbf{z}^*_t \sim \mathcal{N}(\rho, \mathbf{I})$.
For a fair comparison, this transformation also satisfies the condition of $\mathcal{D}(p(\mathbf{z}_s), p(\mathbf{z}^*_t)) \leq \rho$. Then, the image $\mathbf{x}^*_t$ is generated via the diffusion model with $\mathbf{z}^*_t \sim \mathcal{N}(\rho, \mathbf{I})$ and the original prompt $\mathbf{p}_s$. 
As shown in Table \ref{tab:transport_abl}, our transport transformation $T$ significantly outperforms the random transformation. Indeed, using the random transformation even downgrades the performance of the CLIP model since the generation of the diffusion model by using random transformation is uncontrolled.
Meanwhile, by controlling the latent variable $\mathbf{z}^*_t$ via $\mathbf{z}_s$ and $\rho$, as defined in Eqn. \eqref{eqn:transform_function}, the generation of adversarial samples is oriented and significantly improves the CLIP's performance.

\begin{wraptable}[7]{r}{0.4\textwidth}
\centering
\vspace{-5mm}
\caption{The Effectiveness of Pre-trained Latent Diffusion Model.}
\label{tab:ldm_ablation}
\vspace{-2mm}
\resizebox{0.4\textwidth}{!}{
\begin{tabular}{c|l|ccc}
\hline
                       &                    & Zeroshot & Linear Prob & Fine-Tune \\
\hline
\multirow{3}{*}{\rotatebox{90}{CC3M}}  & CLIP & 17.10          & 53.50          & 79.50  \\
& Retrained-LDM  & 18.77	& 55.12	& 80.18 \\
                       & Pretrained-LDM    & \textbf{20.33}    & \textbf{56.14}      & \textbf{81.12}     \\
\hline
\multirow{3}{*}{\rotatebox{90}{CC12M}} & CLIP & 36.50          & 69.00          & 82.10 \\
& Retrained-LDM & 38.26	& 71.11	& 83.06 \\
                       & Pretrained-LDM    & \textbf{39.34}    & \textbf{72.12}       & \textbf{84.67}    \\
\hline
\end{tabular}
}
\end{wraptable}

\textbf{Effectiveness of Pre-trained and Re-trained Diffusion Model}
We compared the pre-trained LDM with a re-trained latent diffusion model on CC3M and CC12M.
We only re-train the second stage of the LDM while we adopt the pre-trained VQ-VAE of LDM \cite{rombach2022high} for the first stage.
As shown in Table \ref{tab:ldm_ablation}, the experimental results show that using adversarial samples generated via our transport transformation has significantly improved the performance in both cases of using re-trained and pre-trained LDM.
However, practically, the performance of using the pre-trained diffusion model outperforms re-training the diffusion model on the corresponding dataset.
This is because the pre-trained latent diffusion model was trained on the large-scale dataset and is able to model the data distribution better than the re-trained latent diffusion on the specific datasets. Therefore, using the pre-trained latent diffusion model is beneficial in terms of not only time efficiency but also performance improvement.

\begin{table}[!b]
\centering
\vspace{-6mm}
\caption{The Effectiveness of Our Proposed Approach on Different Datasets and Different Language-Image Pretraining Models.}
\label{tab:ablation_diffent_data_model}
\setlength{\tabcolsep}{3pt}
\resizebox{1.0\textwidth}{!}{
\begin{tabular}{l | ccc | ccc | ccc}
\hline
                       & \multicolumn{3}{c|}{CC3M}                         & \multicolumn{3}{c|}{CC12M}                        & \multicolumn{3}{c}{LAION400M}                    \\
                       & Zeroshot       & Linear Prob    & Fine-Tune      & Zeroshot       & Linear Prob    & Fine-Tune      & Zeroshot       & Linear Prob    & Fine-Tune      \\

\hline

CLIP                   & 17.10          & 53.50          & 79.50          & 36.50          & 69.00          & 82.10          & 67.00          & 78.60          & 84.70          \\
\textbf{Ours + CLIP}   & \textbf{20.33} & \textbf{56.14} & \textbf{81.12} & \textbf{39.34} & \textbf{72.12} & \textbf{84.67} & \textbf{70.11} & \textbf{80.74} & \textbf{86.98} \\
$\Delta$                  & +3.23           & +2.64           & +1.62           & +2.84           & +3.12           & +2.57           & +3.11           & +2.14           & +2.28           \\

\hline

LaCLIP                 & 21.50          & 56.50          & 81.15          & 48.40          & 72.30          & 82.53          & $-$ & $-$ & $-$ \\
\textbf{Ours + LaCLIP} & \textbf{24.12} & \textbf{58.03} & \textbf{83.11} & \textbf{51.16} & \textbf{74.34} & \textbf{84.68} & 
$-$ & $-$ & $-$ \\
$\Delta$                  & +2.62           & +1.53           & +1.95           & +2.76           & +2.04           & +2.15           & $-$ & $-$ & $-$ \\

\hline

SLIP                   & 23.00          & 65.40          & 81.40          & 40.70          & 73.70          & 83.10          & 70.21          & 80.34          & 85.83          \\
\textbf{SLIP+ Our}     & \textbf{26.97} & \textbf{67.60} & \textbf{83.18} & \textbf{43.13} & \textbf{75.58} & \textbf{84.95} & \textbf{72.53} & \textbf{83.21} & \textbf{87.49} \\
$\Delta$                  & +3.97           & +2.20           & +1.78           & +2.43           & 1.88           & +1.85           & +2.33           & +2.87           & +1.67           \\

\hline
\end{tabular}
}
\end{table}

\noindent
\textbf{Effectiveness of Our Domain Generalization on Different Datasets and CLIP-based Models}
Table \ref{tab:ablation_diffent_data_model} illustrates the results of our proposed approach on three datasets at different scales and CLIP-based models, i.e., CLIP \cite{radford2021learning}, LaCLIP \cite{fan2023improving}, and SLIP \cite{mu2022slip}.
The zero-shot classification results have illustrated the generalizability of our proposed approach on different dataset scales. In particular, our proposed approach improves the zero-shot results of CLIP by $+3.23\%$, $+2.48\%$, and $3.11\%$ on CC3M, CC12M, and LAION400M, respectively. 
Further fine-tuning the model via linear probing or end-to-end fine-tuning significantly improves the performance of the CLIP model. 
The results of fine-tuned models on ImageNet achieved $81.12\%$, $84.67\%$, and $86.98\%$ on CC3M, CC12M, and LAION400M, respectively. 
Our proposed approach is effective not only on different datasets but also with different CLIP-based approaches. By further using better CLIP-based training approaches, i.e., LaCLIP or SLIP, the performance of zero-shot results is significantly improved, up to $75.58\%$ trained LAION-400M using SLIP. By further fine-tuning the SLIP model, our proposed approach achieved state-of-the-art performance on ImageNet1K, i.e., $87.49\%$.
The results in Table  \ref{tab:ablation_diffent_data_model} have confirmed the scalability and generalizability of our approach across the training datasets and CLIP-based models.

\vspace{-2mm}
\subsection{Comparisons With State-of-the-Art Approaches}
\vspace{-2mm}

In this section, we present the results of our approach compared with other augmentation and domain generalization approaches, i.e., ADA \cite{volpi2018generalizing}, AdvStyle \cite{zhong2022adversarial}, and Masking Augmentation (FLIP) \cite{li2023scaling}. 

\textbf{Zero-shot Classification}
Table \ref{tab:dg_comparison} compares our approach with other augmentation and domain generalization methods.
Our proposed approach consistently improves the performance of zero-shot classification.
While the masking augmentation generates masked augmented samples, ADA \cite{volpi2018generalizing} and AdvStyle \cite{zhong2022adversarial} generate the adversarial samples via adversarial training. However, the distribution shift in these methods remains limited compared to our diffusion-based approach.
As a result, our proposed approach significantly outperforms other augmentation and domain generalization approaches. In particular, by pre-training on the large-scale LAION400M dataset, our model achieves the state-of-the-art zero-shot classification performance, i.e., $70.11\%$ and $72.53\%$ by using CLIP and SLIP training.
The results have shown our advantages in improving the generalizability of vision-language models against unknown data distributions.

\textbf{Linear-Probing and End-to-end Fine-tuning Classification}
Table \ref{tab:dg_comparison} illustrates the results of our linear probing and fine-tuning experiments.
Similar to the zero-shot classification results, our linear probing and end-to-end fine-tuning results consistently improve the performance of CLIP \cite{radford2021learning} and SLIP \cite{mu2022slip} and outperform other augmentation approaches.
By pre-training on LAION-400M and further fine-tuning on ImageNet-1K, our training approach achieved state-of-the-art performance, with the accuracy of CLIP and SLIP improved to $86.98\%$ and $87.49\%$.
These results have further confirmed the effectiveness of our approach across evaluation settings and pre-training datasets.

\begin{table}[t]
\centering
\caption{The Comparison With Other Augmentation and Generalization Approaches.}
\label{tab:dg_comparison}
\setlength{\tabcolsep}{3pt}
\resizebox{1.0\textwidth}{!}{
\begin{tabular}{l|ccc|ccc|ccc}
\hline
                & \multicolumn{3}{c|}{CC3M}           & \multicolumn{3}{c|}{CC12M}          & \multicolumn{3}{c}{LAION400M}     \\
                & Zeroshot & Linear Prob & Fine-Tune & Zeroshot & Linear Prob & Fine-Tune & Zeroshot & Linear Prob & Fine-Tune \\
\hline
CLIP            & 17.10     & 53.50       & 79.50     & 36.50    & 69.00       & 82.10     & 67.00    & 78.60       & 84.70     \\
CLIP + Masking  & 17.69 & 54.13 & 80.08 & 37.34 & 70.56 & 82.28 & 68.06 & 78.95 & 85.03 \\
CLIP + ADA      & 18.36 & 55.75 & 80.43 & 38.10 & 70.95 & 82.93 & 68.59 & 79.54 & 85.23 \\
CLIP + AdvStyle & 19.01 & 55.55 & 80.40 & 38.77 & 71.22 & 81.21 & 69.47 & 79.90 & 85.57 \\
CLIP + Ours     & \textbf{20.33} & \textbf{56.14} & \textbf{81.12} & \textbf{39.34} & \textbf{72.12} & \textbf{84.67} & \textbf{70.11} & \textbf{80.74} & \textbf{86.98} \\
\hline

SLIP                 & 23.00          & 65.40          & 81.40          & 40.70          & 73.70          & 83.10          & 70.21          & 80.34          & 85.83          \\
SLIP + Masking       & 24.13          & 65.98          & 81.91          & 41.01          & 73.97          & 83.29          & 70.47          & 80.92          & 86.05          \\
SLIP + ADA           & 24.89          & 66.26          & 82.09          & 41.64          & 74.02          & 83.56          & 70.95          & 81.49          & 86.23          \\
SLIP + AdvStyle      & 25.50          & 66.55          & 82.59          & 42.30          & 74.46          & 84.01          & 71.37          & 81.74          & 86.58          \\
\textbf{SLIP + Ours} & \textbf{26.97} & \textbf{67.60} & \textbf{83.18} & \textbf{43.13} & \textbf{75.58} & \textbf{84.95} & \textbf{72.53} & \textbf{83.21} & \textbf{87.49} \\
\hline
\end{tabular}
}
\vspace{-6mm}
\end{table}

\begin{wraptable}[7]{r}{0.5\textwidth}
\centering
\vspace{-5mm}
\caption{Zero-shot Classification Results on Six Benchmarks, i.e., STL-10, Country-211, Caltech-101, Flowers, Pets, and SUN-397.}
\label{tab:zeroshot_other}
\vspace{-2mm}
\setlength{\tabcolsep}{2pt}
\resizebox{0.5\textwidth}{!}{
\begin{tabular}{l|cccccc}
\hline
                      & \rotatebox{0}{STL-10}         & \rotatebox{0}{Coun-211}     & \rotatebox{0}{Cal-101}    & \rotatebox{0}{Flowers}        & \rotatebox{0}{Pets}           & \rotatebox{0}{SUN-397}         \\
\hline
CLIP                  & 97.30          & 17.80          & 91.20          & 63.90          & 90.10          & 66.80          \\
\textbf{CLIP + Our}   & \textbf{97.58} & \textbf{18.34} & \textbf{93.14} & \textbf{77.12} & \textbf{91.74} & \textbf{68.85} \\
\hline
SLIP & 97.50 & 19.90 & 92.10 & 75.62 & 91.00 & 67.40 \\
\textbf{SLIP + Our} & \textbf{98.87} & \textbf{21.73} & \textbf{94.63} & \textbf{81.35} & \textbf{94.67} & \textbf{70.41} \\
\hline

\end{tabular}
}
\end{wraptable}

\textbf{Other Zero-shot Classification Benchmarks} Table \ref{tab:zeroshot_other} illustrates the results of our proposed approach (pretrained on LAION400M)
on six different zero-shot benchmarks. 
Our approach consistently improves the performance of CLIP and SLIP on all zero-shot classification benchmarks which have illustrated the generalizability of our approach to unseen domains. 
Thanks to our generalization approach, the vision-language foundation model is able to learn better visual representation against the data distribution shift. 
Therefore, the vision-language model can be later well generalized to various downstream tasks.

\vspace{-4mm}
\section{Conclusions, Limitations, and Broader Impact}
\vspace{-2mm}

\textbf{Conclusions:}
This paper has introduced a novel, simple yet efficient diffusion-based domain generalization approach to the vision-language foundation model. Under our theoretical analysis, we introduced a new efficient sampling to generate new diffusion-based adversarial samples based on our proposed transport transformation to improve the generalizability of the vision-language foundation model. Our experimental results on various benchmarks have illustrated the effectiveness of our generalization approach to the vision-language foundation model.

\textbf{Limitations:}
Our paper has chosen specific network configurations and learning hyperparameters to support our hypothesis. However, the other aspects of learning have not been fully investigated due to hardware constraints, e.g., the larger network sizes, the different diffusion models, the larger scale of pre-training datasets, etc.
Additionally, the larger pre-training dataset may require larger computational time to generate diffusion-based adversarial samples.

\textbf{Broader Impacts:}
Our paper studies the problem of domain generalization, which is a step forward in improving the generalizability of the vision-language foundation model.
Our contributions have emphasized the relationship between diffusion and domain generalization, which can later be used to improve the performance of vision-language models.
Our approach helps to increase the robustness of foundation models across various zero-shot downstream tasks.

\bibliographystyle{abbrv}
\bibliography{references}

\newpage

\begin{center}
    \large \textbf{Appendix}
\end{center}

\setcounter{section}{0}

\section{Proof of Proposition 1}

\textbf{\textit{Proposition 1:}} \textit{Given $\mathbf{z}_s \in \mathcal{N}(\mathbf{0, I})$ and $\alpha$ ($-\rho \leq \alpha \leq \rho$), the condition of distance between distributions $\mathcal{D}(p_s(\mathbf{z}_s), p_t(\mathbf{z}^*_t)) \leq \rho$ holds if the transport transformation $\mathcal{T}$ is defined as $\mathcal{T}(\mathbf{z}_s, \rho) = \frac{\mathbf{z}_s + \mathcal{N}(\alpha\sqrt{2}\mathbf{I}, \mathbf{I})}{\sqrt{2}}$.}

\textbf{Proof:} The proposition can be sufficiently proven via the Wasserstein distance between two distributions. As the latent variable $\mathbf{z}_s$ belong to the Normal distribution, i.e., $\mathbf{z}_s \in \mathcal{N}(\mathbf{0, I})$, the transformed latent variable $\mathbf{z}^*_t$ via the transformation $\mathcal{T}$ should belong to $\mathbf{z}^*_t = \frac{\mathbf{z}_s + \mathcal{N}(\alpha\sqrt{2}\mathbf{I}, \mathbf{I})}{\sqrt{2}} \sim \mathcal{N}(\alpha, \mathbf{I})$. Then, the transportation cost between two distributions $p_s$ and $p_t$ measured via the Wasserstein distance can be defined as follows:
\begin{equation}\label{eqn:distance_w}
\begin{split}
    \mathcal{D}(\mathcal{N}(\boldsymbol{\mu}_s, \boldsymbol{\Sigma}_s), \mathcal{N}(\boldsymbol{\mu}_t, \boldsymbol{\Sigma}_t)) = || \boldsymbol{\mu}_s - \boldsymbol{\mu}_t ||^2_2 +  \operatorname{tr}\left[\boldsymbol{\Sigma}_s+\boldsymbol{\Sigma}_t-2(\boldsymbol{\Sigma}_s^{1/2}\boldsymbol{\Sigma}_t\boldsymbol{\Sigma}_1^{1/2})^{1/2}\right].
\end{split}
\end{equation}
Since $\boldsymbol{\mu}_s = \mathbf{0}$, $\boldsymbol{\mu}_t = \alpha$, the $\boldsymbol{\Sigma}_s$ and $\boldsymbol{\Sigma}_t$ of the two data distributions is $\mathbf{I}$, the distance between two data distribution defined in Eqn. \eqref{eqn:distance_w} can be rewritten as:
\begin{equation}
\begin{split}
    \mathcal{D}(\mathcal{N}(\boldsymbol{\mu}_s, \boldsymbol{\Sigma}_s), \mathcal{N}(\boldsymbol{\mu}_t, \boldsymbol{\Sigma}_t)) = ||\mathbf{0} - \alpha||^2_2  = \alpha &\leq \rho \\
    \Rightarrow \quad \mathcal{D}(p_s(\mathbf{z}_s), p_t(\mathbf{z}^*_t)) &\leq \rho \quad\quad \text{(Q.E.D)}
\end{split}
\end{equation}

\section{Additional Ablation Study}

\begin{wraptable}[10]{r}{0.4\textwidth}
\centering
\caption{The Effectiveness of Vision Network Backbone.}
\label{tab:backbone_ablation}
\resizebox{0.4\textwidth}{!}{
    \begin{tabular}{l|l|ccc}
    \hline
                            & Different Work                      & Zeroshot & Linear Prob & Fine-Tune \\
    \hline
                            & RN50                                & 16.13    & 50.02       & 78.12     \\
                            & RN50  + Our                         & \textbf{18.87}    & \textbf{52.11}       & \textbf{80.08}     \\
                            \cdashline{2-5}
                            & ViT-B-16                            & 17.10    & 53.50       & 79.50     \\
    \multirow{-4}{*}{\rotatebox{90}{CC3M}}  & ViT-B-16 + Our                      & \textbf{20.33}    & \textbf{56.14}       & \textbf{81.12}     \\
    \hline
                            & RN50        & 35.02    & 67.42       & 80.81     \\
                            & RN50  + Our & \textbf{37.05}    & \textbf{70.21}       & \textbf{82.36}     \\
                            \cdashline{2-5}
                            & ViT-B-16                            & 36.50    & 69.00       & 82.10     \\
    \multirow{-4}{*}{\rotatebox{90}{CC12M}} & ViT-B-16 + Our                      & \textbf{39.34}    & \textbf{72.12}       & \textbf{84.67}    \\
    \hline
    \end{tabular}
}
\end{wraptable}
\textbf{Effectiveness of Different Backbone}
The results in Table \ref{tab:backbone_ablation} illustrate the effectiveness of our approach in different backbones, i.e., ResNet-50 (RN50) and ViT-B/16.
By using the better backbone, the performance of the CLIP model is major improved.
In particular, on the zero-shot classification benchmarks, the performance of our approach trained on CC3M and CC12M is improved from $18.77\%$ to $20.33\%$ and from $37.05\%$ to $39.34\%$, respectively.
The performance is even further majorly improved in both backbones with further fine-tuning.

\section{Discussion of Limitations and Broader Impact}

\textbf{Limitations}
In our paper, we opt for a particular set of hyperparameters and learning methodologies to bolster our hypothesis.
Although our proposed approach has shown its effectiveness in improving the generalizability of the vision-language foundation model, it could potentially consist of several limitations.
First, the choice of different contrastive learning losses in the vision-language models should be exploited.
Second, the different visual and textual encoders have not been fully investigated in our study.
Besides, the larger number of DDIM steps in the data generation process of image generation should be studied in future work.
Third, the data generation process via the diffusion model requires high computational resources and a large amount of time.
Additionally, in our paper, we only consider the images conditioned on the text prompts. However, the different conditions, e.g., image or object layouts, semantic segmentation, etc., should be considered in future studies. 
These constraints may motivate new research studies to enhance the diffusion-based domain generalization approach to vision-language foundation models.

\textbf{Other Potential Social Broader Impacts}
Our paper has introduced a novel diffusion-based approach to domain generalization in the vision-language foundation model.
Our approach has improved the performance of the foundation model on various downstream tasks.
However, we acknowledge that the large-scale diffusion model, i.e., LDM \cite{rombach2022high}, trained on extreme-scale data could potentially produce inappropriate images and even hallucinations.
Thus, the vision-language models could accidentally learn these pieces of information.
In addition, since the data generation process requires high computational resources and a large amount of time, it could potentially produce a higher carbon footprint.

\section{Adversarial Samples}

\begin{figure}[!b]
    \centering
    \includegraphics[width=0.9\textwidth]{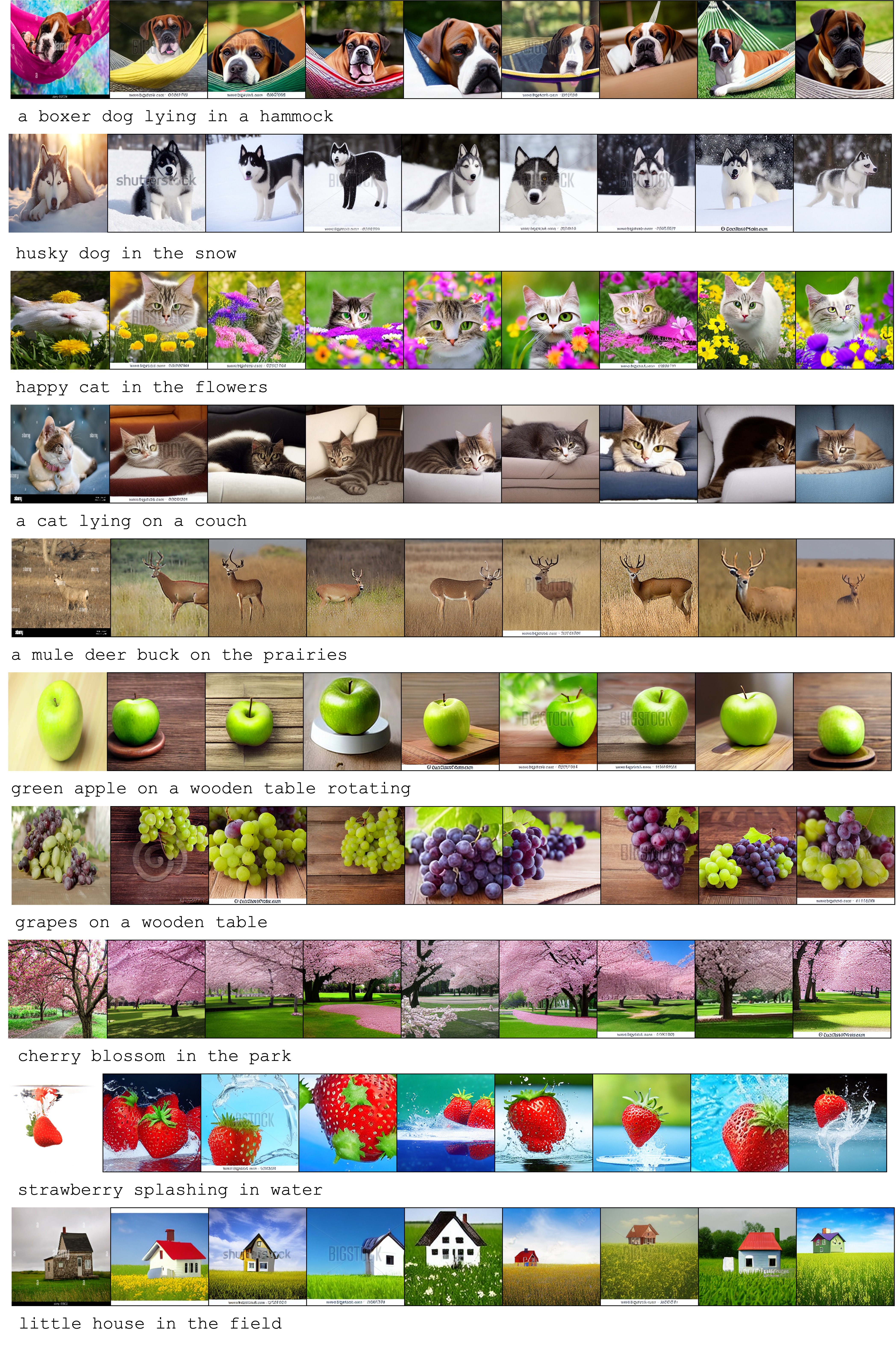}
    \caption{Our Diffusion-based Adversarial Samples. The first figure of each row is the original image.}
    \label{fig:more_adv_sample}
\end{figure}

Fig. \ref{fig:more_adv_sample} illustrates our diffusion-based adversarial samples generated via our proposed transport transformation with the latent diffusion model \cite{rombach2022high}.

\section{Qualitative Results of Zeroshot Predictions}

\begin{figure}[H]
    \centering
    \includegraphics[width=1.0\textwidth]{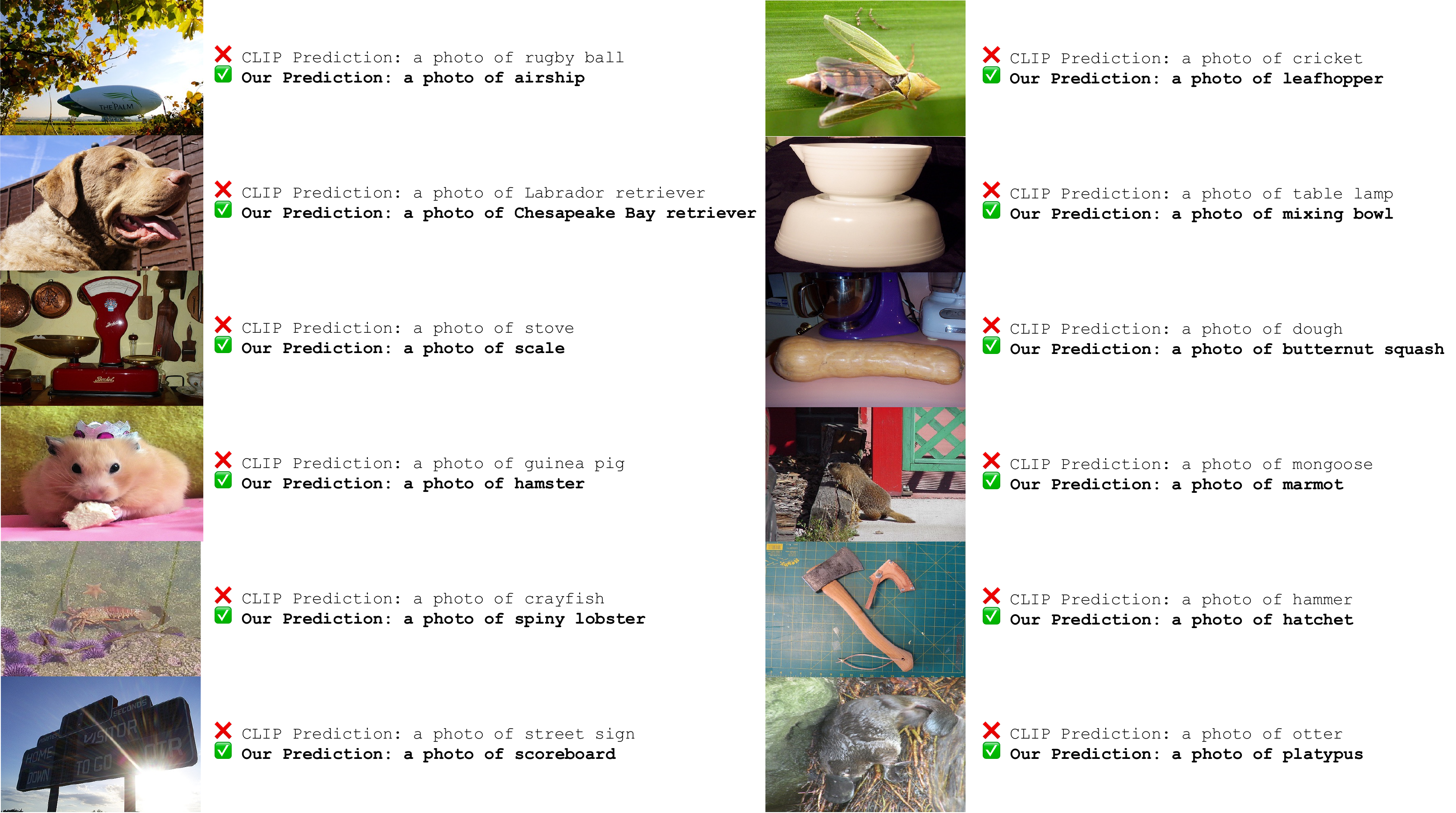}
    \caption{Our Zeroshot Predictions on ImageNet-1K.}
    \label{fig:zero_shot_pred}
\end{figure}

Fig. \ref{fig:zero_shot_pred} visualizes our zero-shot predictions on ImageNet-1K compared to CLIP trained on LAION400M.

\end{document}